\documentclass{article}

    \PassOptionsToPackage{numbers}{natbib}

 \usepackage[main, final]{neurips_2026}

\usepackage[utf8]{inputenc} 
\usepackage[T1]{fontenc}    
\usepackage{hyperref}       
\usepackage{url}            
\usepackage{booktabs}       
\usepackage{amsfonts}       
\usepackage{nicefrac}       
\usepackage{microtype}      
\usepackage{xcolor}         
\usepackage{amsmath}
\usepackage{graphicx}
\usepackage{algorithm}
\usepackage{algpseudocode}
\usepackage{subcaption}
\usepackage{float}
\usepackage{bbm}

\title{Action Chunking Proximal Policy Optimization with Feedback Correction}

\author{%
  Sanghyun Hahn\thanks{Work done at Seoul National University} \\
  Cornell University \\
  \texttt{steve0221@cs.cornell.edu} \\
  \And
  Jonghyun Choi \\
  Seoul National University \\
  \texttt{jonghyunchoi@snu.ac.kr} \\
}

\begin{document}
\maketitle

\begin{abstract}
Action chunking provides temporal abstraction in reinforcement learning by selecting short action sequences instead of individual actions, but many existing approaches face two limitations in high-dimensional robotic control. First, many rely on value functions over action chunks, which can be difficult to learn as action dimensionality and chunk length grow. Second, executing chunks open-loop removes within-chunk feedback, limiting reactivity in contact-rich tasks. We present Action Chunking PPO (ACPPO), a PPO extension that uses a chunked actor while retaining a standard state-value critic, thereby avoiding chunked Q-functions. We further propose ACPPO-Corr, which augments the chunk planner with a stepwise feedback corrector that adjusts planned actions online within each chunk. Across 25 simulated robotics tasks from IsaacGym and Bi-DexHands, spanning locomotion, arm manipulation, and dexterous hand-object interaction, ACPPO-Corr achieves the strongest aggregate performance among evaluated methods and performs best on both decision-frequency-sensitive and decision-frequency-neutral task subsets. Ablations show that moderate chunk lengths work best and that corrector regularization is important for balancing chunk-level planning with local feedback. These results suggest that action chunking can be effective in online PPO when chunk-level planning is paired with closed-loop correction.
The code is available at: \url{https://github.com/hshhahn/ACPPO.git}
\end{abstract}

\section{Introduction}
	
Action Chunking~\citep{zhaolearning, bharadhwaj2024roboagent, chi2025diffusion, black2024pi0, kim2025fine, lee2025molmoact} has been a popular design choice in robot imitation learning, where the policy predicts a sequence of actions instead of a single action at each decision step. 
This reduces the effective control horizon at inference, mitigates compounding errors in the continuous action space~\citep{simchowitz2025pitfalls, zhang2025action}, and captures non-Markovian behaviors inherent in human demonstrations~\citep{zhaolearning}.

Recent work has incorporated action chunking into reinforcement learning (RL), treating action chunks as temporally extended actions in a semi-Markov decision process~\citep{sutton1999between, li2025reinforcement, wang2025vla, park2026scalable, yang2025ac3, nagy2026sear, kim2026deas, li2026decoupled} or exploiting chunked critics~\citep{seo2025coarsetofine, litop, tian2026chunking}. 
Compared to stepwise control, action chunking RL can offer several advantages. 
First, multi-step actions shorten the decision horizon and accelerate reward propagation, which is beneficial in long-horizon problems~\citep{li2025reinforcement, park2026scalable, kim2026deas}. 
Second, chunked policies allow temporally coherent exploration, which can lead to deeper state-space coverage and improved exploration efficiency~\citep{li2025reinforcement, nagy2026sear, wang2025vla}. 
 
However, these benefits come with two major challenges when applied to high-dimensional robotic control: chunked action-value function learning and the loss of within-chunk reactivity.
First, most prior action-chunking RL methods rely on chunked Q-functions, $Q(s_t,a_{t:t+h-1})$, requiring the critic to evaluate high-dimensional action sequences.
Prior work shows that this can make value learning more difficult, cause overestimation and instability, and make policy extraction increasingly challenging as the chunk length grows~\citep{seo2025coarsetofine, kim2026deas, li2026decoupled}.
As a result, some prior approaches rely on offline pre-training~\citep{li2025reinforcement, kim2026deas, seo2025coarsetofine} or use heavier architectures, such as transformers, to model long action sequences~\citep{nagy2026sear, litop}.
This limitation is particularly important in our setting, where the policy must operate in high-dimensional environments, and obtaining a reliable offline dataset is challenging.

Second, executing a predicted action chunk in an open-loop manner causes the loss of within-chunk reactivity: after the action sequence has been selected, the actions cannot adapt to environmental stochasticity, contact changes, or unexpected perturbations.
This loss of within-chunk feedback is especially problematic in contact-rich robotic manipulation, including dexterous hand control~\citep{johansson2009coding}, where contact dynamics can change rapidly and successful control often depends on timely closed-loop correction~\citep{xue2025reactive, zheng2026omnivtavisuotactileworldmodeling}.

In this work, we propose Action Chunking PPO (ACPPO), an extension of PPO that brings action chunking into a fully online, on-policy RL setting while keeping a standard state-value critic. 
ACPPO introduces temporal abstraction to the actor through a chunked planner that predicts short action sequences, while retaining the state-only critic of PPO~\citep{schulman2017proximal}. 
The critic itself is not chunked: it remains a standard state-value function $V(s)$, while the PPO surrogate uses a chunk-level advantage computed from an $h$-step chunk return. 
Thus, ACPPO introduces multi-step actor credit assignment without learning a chunked Q-function.

Building on this formulation, we propose ACPPO-Corr, which adds a stepwise feedback corrector that adjusts the planned actions every step. 
This combines low-frequency chunk planning with high-frequency closed-loop control, recovering within-chunk reactivity while maintaining the benefits of action chunking. 
Crucially, instead of relying on prior demonstrations, offline pre-training, or a frozen pretrained chunked policy, we jointly train the planner and corrector within a single fully online, on-policy RL framework.
Empirically, ACPPO-Corr achieves the strongest performance across 25 simulated robotics tasks from IsaacGym~\citep{makoviychuk2021isaac} and Bi-DexHands~\citep{chen2022towards}, including both decision-frequency-sensitive and decision-frequency-neutral subsets.

\section{Related Work}

\paragraph{Action Chunking in Reinforcement Learning.}
Multi-step returns have been a popular choice in RL for trading off the bias of one-step TD targets against the variance of Monte Carlo returns~\citep{sutton1988learning}. 
Recent action chunking RL methods extend this idea to temporally extended control by learning over short action sequences, often through chunked critics or action-sequence policies. 
One line of work focuses on critic-side temporal abstraction while keeping a stepwise policy or omitting an explicit actor~\citep{seo2025coarsetofine,tian2026chunking,song2026chunk,litop}. 
These methods move temporal abstraction to the critic by learning values over short action sequences, reducing the compounding errors from the bootstrapped value estimates~\citep{song2026chunk}. 
A second line of work jointly learns chunked actors and chunked critics in offline, offline-to-online, or online settings~\citep{park2026scalable,li2025reinforcement,kim2026deas,yang2025ac3,nagy2026sear}. 
These methods benefit from chunked actors, which produce temporally coherent action sequences and can encourage coherent exploration. 
Despite differences in their settings, a large fraction of these methods rely on chunked Q-functions and are based on off-policy reinforcement learning methods, often augmented with heavier architectures, such as transformers. 
This limitation is particularly important in our setting:
in high-dimensional continuous control, collecting expert demonstrations is costly, whereas parallel simulators make online rollouts practical~\citep{makoviychuk2021isaac, rudin2022learning}. 
Consequently, PPO-type on-policy methods remain strong baselines on high-DoF manipulation benchmarks~\citep{rajeswaran2018learning, chen2022towards}. 
Our work studies action chunking in a fully online, on-policy PPO setting with a chunked actor and a standard state-value critic.
This differs both from online chunked actor-critic methods based on chunked Q-functions~\citep{nagy2026sear, yang2025ac3}, and from VLA post-training that applies PPO-style updates to pretrained action-chunking policies while relying on expert demonstrations~\citep{wang2025vla}.
In contrast to chunked Q-learning, the critic input in ACPPO does not grow with the chunk length: the critic remains $V_\phi(s)$, and chunking is applied by the actor and the chunked advantage rather than through a chunked Q-function $Q(s, a_{t:t+h-1})$.

\paragraph{Residual correction and reactive execution.}
A related line of work addresses the loss of reactivity or execution mismatch in chunked policies, either by learning corrective residuals or by modifying real-time inference.
Classical residual RL augments a frozen controller with a learned corrective policy~\citep{johannink2019residual}.
In robot learning from demonstrations,~\citet{ankile2025imitation} learn a closed-loop residual policy on top of a frozen behavior-cloning chunked planner, and~\citet{ankile2025residual} study off-policy residual fine-tuning of such frozen base policies on high-DoF robots.
For pretrained chunked policies,~\citet{xue2025reactive} combine a slow chunk-level planner with a fast tactile controller, while~\citet{liubidirectional} improve reactivity through test-time closed-loop resampling of chunk predictions.
For real-time deployment of pretrained chunked policies,~\citet{black2025real} generate the next chunk while executing the current one,~\citet{sendai2025leave} add a per-step correction head to off-the-shelf VLA action chunks, and~\citet{wang2026real} learn corrective adjustments on top of a pretrained policy through masked action chunking.
ACPPO-Corr is closest to this line of work, but differs in the learning setup: instead of correcting a pretrained, frozen chunked policy, we jointly train a chunk planner and a stepwise corrector within a fully online, on-policy RL formulation.
Therefore, the corrector is not an add-on execution module, but part of the policy optimized jointly with the chunk planner from scratch.

\section{Preliminaries}

\paragraph{Problem setup.} We consider a Markov decision process (MDP)~\citep{sutton1998reinforcement} $\mathcal{M} = (\mathcal{S}, \mathcal{A}, P, r, \gamma)$, where $s_t \in \mathcal{S}$ is the state, $a_t \in \mathcal{A}$ is the action, $r_t = r(s_t, a_t)$ is the reward, and $P(s_{t+1} \mid s_t, a_t)$ is the transition kernel. The behavior of the agent is determined by a stochastic policy $\pi_\theta(a_t \mid s_t)$, and the objective is to maximize the expected discounted return
\begin{equation*}
J(\pi_\theta)
=
\mathbb{E}_{\tau \sim \pi_\theta}
\left[
\sum_{t=0}^{\infty} \gamma^t r_t
\right].
\end{equation*}
For continuous action spaces, we assume $\mathcal A \subset \mathbb{R}^{d_a}$ and parameterize the policy as a diagonal Gaussian.

\paragraph{Proximal Policy Optimization.}
Given a batch of trajectories collected by a policy $\pi_{\theta_{\mathrm{old}}}$, Proximal Policy Optimization (PPO)~\citep{schulman2017proximal} performs multiple epochs of first-order updates on a clipped surrogate objective that controls the deviation of $\pi_\theta$ from $\pi_{\theta_{\mathrm{old}}}$. Defining the importance ratio
\begin{equation*}
\rho_t(\theta)
=
\frac{\pi_\theta(a_t \mid s_t)}
{\pi_{\theta_{\mathrm{old}}}(a_t \mid s_t)},
\end{equation*}
PPO maximizes the surrogate:
\begin{equation*}
J_{\mathrm{PPO}}(\theta)
=
\mathbb{E}
\left[
\min\!\left(
\rho_t(\theta) \hat{A_t},\;
\mathrm{clip}(\rho_t(\theta), 1-\epsilon, 1+\epsilon) \hat{A_t}
\right)
\right],
\end{equation*}
where $\hat{A_t}$ is the advantage estimate. 
To compute this advantage, a state-value critic $V_\phi(s_t)$ is trained jointly with the actor.
We define the one-step temporal-difference residual as
\begin{equation*}
\delta_t
=
r_t + \gamma  V_\phi(s_{t+1}) - V_\phi(s_t).
\end{equation*}
$\hat{A_t}$ is obtained by the Generalized Advantage Estimate (GAE)~\citep{schulman2015high}
\begin{equation*}
\hat{A_t} = 
\hat A_t^{\mathrm{GAE}}
=
\delta_t + \gamma \lambda \hat A_{t+1}^{\mathrm{GAE}}.
\end{equation*}
Finally, the corresponding value target to train the critic is
\begin{equation*}
\hat{R_t} = \hat A_t^{\mathrm{GAE}} + V_\phi(s_t).
\end{equation*}

\paragraph{Multi-step returns and control.}
An $h$-step bootstrapped return aggregates rewards over a short horizon:
\begin{equation*}
G_t^{(h)}
=
\sum_{j=0}^{h-1} \gamma^j r_{t+j}
\;+\;
\gamma^h V_\phi(s_{t+h}).
\end{equation*}
These multi-step returns interpolate between one-step TD targets and Monte Carlo returns, yielding a bias-variance tradeoff. 
In our setting, a chunk-level planner (chunked actor) predicts an open-loop action sequence of length $h$
\begin{equation*}
\mathbf{u}_{t_0:t_0+h-1}
=
[{u}_{t_0,0}, \dots, {u}_{t_0,h-1}]
\in \mathbb{R}^{h \times d_a},
\end{equation*}
where ${u}_{t_0,k}$ denotes the action associated with the $k$-th step of the chunk.
We use $u$ for planned chunk actions and $a$ for the action executed in the environment.

\section{Method}
\begin{figure}
    \centering
    \includegraphics[width=1.0\linewidth]{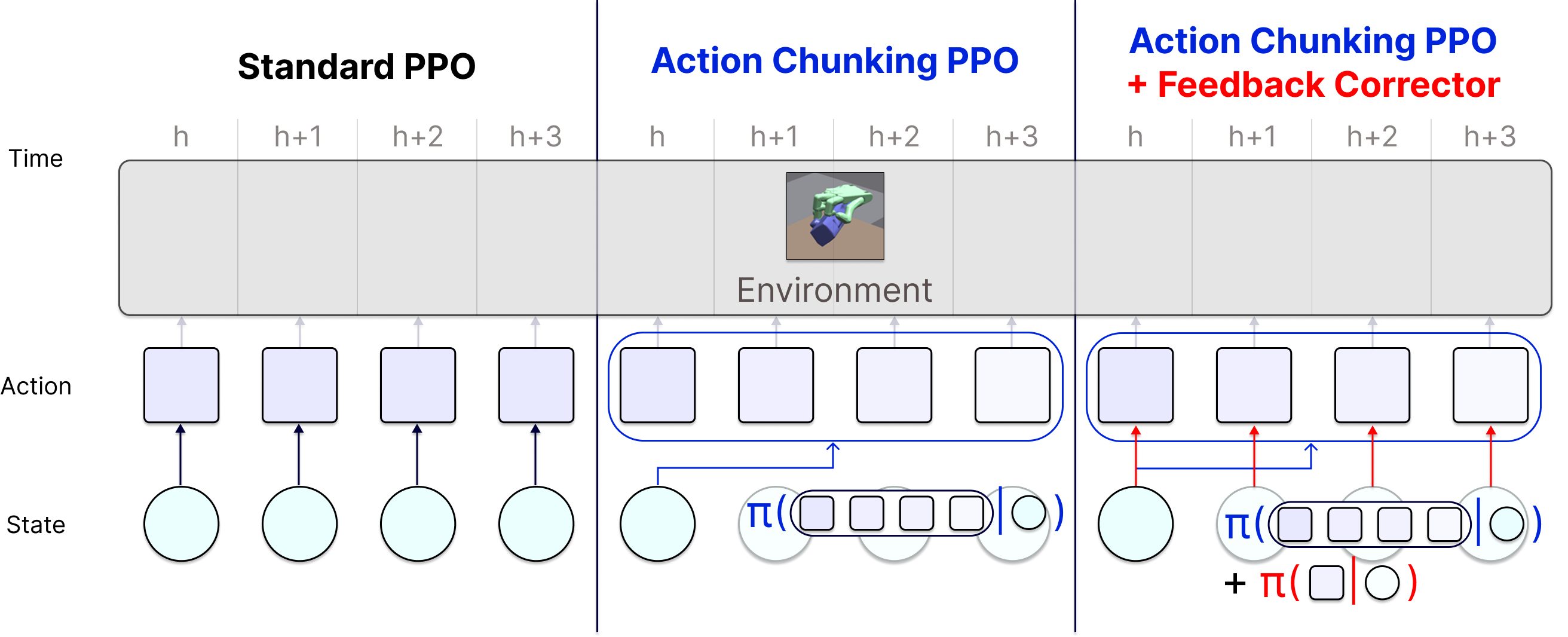}
    \caption{Stepwise comparison between PPO, ACPPO, and ACPPO-Corr. PPO runs the policy closed-loop. ACPPO runs the policy open-loop. ACPPO-Corr corrects the open-loop policy with a closed-loop feedback corrector. In ACPPO-Corr, the open-loop policy and the closed-loop corrector are jointly trained.}
    \label{fig:PPO/ACPPO/ACPPO-Corr}
\end{figure}

We first introduce \textbf{Action Chunking PPO (ACPPO)}, which brings action chunking into PPO while keeping a standard state-value critic. 
This avoids chunked Q-functions and keeps the learning setup fully online and on-policy. 
We then extend ACPPO to \textbf{ACPPO-Corr} by adding a stepwise feedback corrector for tasks where within-chunk reactivity is important.

\subsection{Action Chunking PPO (ACPPO)}
For any step $t$, let
\begin{equation*}
t_0 = \left\lfloor \frac{t}{h} \right\rfloor h,
\qquad
k = t - t_0 \in \{0,\dots,h-1\},
\end{equation*}
where $t_0$ is the start of the current chunk and $k$ is the within-chunk index. 
At chunk boundaries, a chunked planner $\pi^{(h)}_{\theta_{\mathrm{pl}}}$ predicts an $h$ step action sequence
\begin{equation*}
\mathbf{u}_{t_0:t_0+h-1}
=
\pi^{(h)}_{\theta_{\mathrm{pl}}}(s_{t_0})
=
[u_{t_0,0}, \dots, u_{t_0,h-1}]
\in \mathbb{R}^{h \times d_a}.
\end{equation*}
At step $t=t_0+k$, ACPPO uses the $k$-th planned action $u_{t_0,k}$ as the policy mean and samples the executed action $a_t$ from a Gaussian. 
In this way, temporal abstraction is introduced at the actor side, while the critic remains state only. 
Accordingly, we also form the advantage and the importance sampling ratio at the chunk level.
The critic is trained with the standard PPO value loss using stepwise GAE returns.

\paragraph{Chunked advantage.}
Instead of using separate advantages at each timestep, ACPPO calculates a single chunked advantage at the chunk-start~\citep{sutton1999between}.
We define the $h$-step chunked advantage as
\begin{equation*}
A_h^\pi(s_{t_0}, \mathbf{a}_{t_0:t_0+h-1})
:=
\mathbb{E}\!\left[
\sum_{j=0}^{h-1} \gamma^j r_{t_0+j}+ \gamma^h V^\pi(s_{t_0+h}) - V^\pi(s_{t_0})
\;\middle|\;
s_{t_0}, \mathbf{a}_{t_0:t_0+h-1}
\right].
\label{eq:macro_adv_exact}
\end{equation*}
In practice, we estimate it as a combination of the $h$-step chunk head and a standard GAE tail at the next chunk boundary:
\begin{equation}
\hat A_{t_0}^{\mathrm{chunk}}
=
\sum_{j=0}^{h-1} \gamma^j \delta_{t_0+j}
+
\gamma^h \hat A_{t_0+h}^{\mathrm{GAE}},
\label{eq:chunk_gae_impl}
\end{equation}
where $\delta_t := r_t + \gamma V_\phi(s_{t+1}) - V_\phi(s_t)$ is the stepwise residual and $\hat A_t^{\mathrm{GAE}}$ is the standard stepwise GAE.
Appendix~\ref{app:chunked_advantage_bias} shows that the $h$-step head telescopes intermediate value terms with a bias--variance tradeoff as the chunk length increases.

\paragraph{Projected state critic.}
\label{sec:projected_state_critic}
When executing chunked actions, the exact value of a mid-chunk step may
depend on active chunk variables $z_t$ such as the chunk-start state, planned chunk, and
within-chunk index. 
Let $\widetilde V^\pi(s_t,z_t)$ denote the exact augmented-state value. 
Under a squared value loss, the best state-only predictor is
\[
V_*^\pi(s)
=
\mathbb{E}\!\left[\widetilde V^\pi(s_t,z_t)\mid s_t=s\right],
\]
where the expectation is over chunk variables induced by the current policy. 
Our critic $V_\phi(s)$ approximates this projection. 
In Eq.~\ref{eq:chunk_gae_impl}, the $h$-step actor advantage uses only the boundary evaluations $V_\phi(s_{t_0})$ and $V_\phi(s_{t_0+h})$, with intermediate
value terms telescoping as shown in Appendix~\ref{app:chunked_advantage_bias}.

\paragraph{Chunked importance sampling ratio.}
ACPPO uses the importance ratio of the joint chunk distribution rather than per-step ratios.
Let $\mathcal T_{t_0}\subset\{0,\dots,h-1\}$ denote the valid chunk prefix~\citep{black2025training} before any mid-chunk episode termination.
\begin{equation*}
\rho_{t_0}^{(h)}(\theta)
=
\frac{
\pi_{\theta}^{(h)}(\mathbf{a}_{t_0:t_0+h-1} \mid s_{t_0})
}{
\pi_{\theta_{\mathrm{old}}}^{(h)}(\mathbf{a}_{t_0:t_0+h-1} \mid s_{t_0})
}
=
\prod_{k\in\mathcal T_{t_0}}
\frac{
\pi_{\theta,k}^{(h)}(a_{t_0+k}\mid s_{t_0})
}{
\pi_{\theta_{\mathrm{old}},k}^{(h)}(a_{t_0+k}\mid s_{t_0})
}.
\end{equation*}
The clipped chunk-level surrogate becomes
\begin{equation*}
J_{\mathrm{chunk}}(\theta)
=
\mathbb{E}_{t_0}
\left[
\min\!\left(
\rho_{t_0}^{(h)}(\theta)\hat A_{t_0}^{\mathrm{chunk}},
\;
\mathrm{clip}\!\left(\rho_{t_0}^{(h)}(\theta),\,1-\epsilon,\,1+\epsilon\right)\hat A_{t_0}^{\mathrm{chunk}}
\right)
\right].
\label{eq:chunked_surrogate}
\end{equation*}
If the action at offset $j$ terminates the episode, then $\mathcal T_{t_0}=\{0,\dots,j\}$ and the remaining offsets are masked.
In vectorized rollouts, a mid-chunk terminated environment is held as inactive padding until the next global chunk boundary, where a new planner is called.

\begin{algorithm}[t]
\caption{ACPPO-Corr}
\label{alg:acppo_corr}
\begin{algorithmic}[1]
\Require planner $\pi_{\theta_{\mathrm{pl}}}$, corrector/value network $\pi_{\theta_{\mathrm{co}},\phi}$, chunk length $h$, rollout horizon $T$
\For{$e=1,\dots,N$}
    \State $\textsc{freeze}_{\rm pl} \gets \mathbbm{1}\{e \leq N_{\rm warm}\}$
    \For{$t=0,\dots,T-1$}
        \If{$t \bmod h = 0$}
            \State $t_0 \gets t$, \quad $\mathbf{u}_{t_0:t_0+h-1} \gets \pi_{\theta_{\mathrm{pl}}}(s_{t_0})$
            \Comment{plan an action chunk}
        \EndIf
        \State $k \gets t-t_0$
        \State $(c_t,\sigma_t,V_t) \gets \pi_{\theta_{\mathrm{co}},\phi}(s_t)$
        \Comment{add closed-loop correction}
        \State Sample $a_t \sim \mathcal{N}(u_{t_0,k}+c_t,\mathrm{diag}(\sigma_t^2))$
        \State Execute $a_t$ and store $(s_t,s_{t_0},a_t,r_t,d_t,\log \pi(a_t),V_t,k)$
    \EndFor
    \State Compute stepwise GAE $\hat A_t^{\mathrm{GAE}}$ and chunked advantages $\hat A_{t_0}^{\mathrm{chunk}}$ using Eq.~\ref{eq:chunk_gae_impl}
    \State For mid-chunk terminations, mask the remaining chunk suffix 
    \For{each PPO minibatch}
        \If{\textbf{not} $\textsc{freeze}_{\rm pl}$}
            \State Evaluate $\mathcal{L}_{\mathrm{total}}$ (Eq.~\ref{eq:acppo_corr_total_loss}) with planner-side stop-grads: $u_{t_0,k}+\mathrm{sg}(c_t),\ \mathrm{sg}(\sigma_t)$
            \State Update $\theta_{\mathrm{pl}}$ with Adam
        \EndIf
        \State Evaluate $\mathcal{L}_{\mathrm{total}}$ using corrector-side stop-gradients: $\mathrm{sg}(u_{t_0,k})+c_t,\ \sigma_t$
        \State Update $(\theta_{\mathrm{co}},\phi)$ with Adam
    \EndFor
\EndFor
\end{algorithmic}
\end{algorithm}

\subsection{ACPPO-Corr: ACPPO With a Stepwise Feedback Corrector}

\paragraph{Adding the stepwise corrector.}
ACPPO addresses the first part of the problem: it brings action chunking into the PPO framework with a standard value critic, avoiding the need to train $Q(s_t,a_{t:t+h-1})$.
However, ACPPO still executes the planned chunk open loop once the chunk has started.
In dexterous manipulation and locomotion, this can be a limitation since contacts and object motion can change quickly~\citep{li2026decoupled, liubidirectional, xue2025reactive}. 
In such settings, low-frequency planning can be useful, but low-frequency reacting is not always sufficient.

To recover within-chunk reactivity, we keep the ACPPO planner for chunk-level means and add a corrector $\pi_{\theta_{\mathrm{co}}}$ that observes the current state at every control step:
\begin{equation*}
({c}_t, {\sigma}_t)
=
\pi_{\theta_{\mathrm{co}}}(s_t).
\end{equation*}
Here, ${c}_t$ is a stepwise correction to the planner mean, while ${\sigma}_t$ is the stepwise exploration scale.
The executed Gaussian policy becomes
\begin{equation*}
\pi_\theta(a_t \mid s_t, s_{t_0}, k)
=
\mathcal{N}\!\left(
{u}_{t_0,k} + {c}_t,\;
\mathrm{diag}({\sigma}_t^2)
\right),
\end{equation*}
where $\theta = (\theta_{\mathrm{pl}}, \theta_{\mathrm{co}})$.
Here, the planner provides chunk-level structure, while the corrector provides stepwise feedback and stepwise exploration inside the chunk.

\paragraph{Corrected importance sampling ratio.}
ACPPO-Corr uses the same chunked advantage in Eq.~\ref{eq:chunk_gae_impl}, but the chunked importance sampling ratio is now also conditioned on the current state. 
For a valid chunk prefix $\mathcal T_{t_0}$~\citep{black2025training}, the importance ratio is
\begin{equation}
\rho_{t_0}^{\mathrm{corr}}(\theta)
=
\prod_{k\in\mathcal T_{t_0}}
\frac{
\pi_\theta(a_{t_0+k}\mid s_{t_0+k},s_{t_0},k)
}{
\pi_{\theta_{\mathrm{old}}}(a_{t_0+k}\mid s_{t_0+k},s_{t_0},k)
}.
\label{eq:acppo_corr_ratio}
\end{equation}
In practice, the adaptive KL scheduling of PPO~\citep{schulman2017proximal} and short chunk lengths keep the empirical clip fraction close to PPO. 
The full diagnostics are reported in Appendix~\ref{app:ratio_diagnostics}.

\paragraph{Decoupled planner and corrector updates.}
We decouple planner and corrector optimization with stop gradients. Let $\mathrm{sg}(\cdot)$ denote stop gradient. For each minibatch, we evaluate the chunk surrogate with two policy parameterizations:
\begin{equation*}
\pi_{\theta}^{m}(a_t \mid s_t,s_{t_0},k)
=
\begin{cases}
\mathcal{N}\!\left(
u_{t_0,k}+\mathrm{sg}(c_t),\;
\mathrm{diag}(\mathrm{sg}(\sigma_t)^2)
\right),
& m=\mathrm{pl},\\[1mm]
\mathcal{N}\!\left(
\mathrm{sg}(u_{t_0,k})+c_t,\;
\mathrm{diag}(\sigma_t^2)
\right),
& m=\mathrm{co}.
\end{cases}
\label{eq:stopgrad_policies}
\end{equation*}
Using the chunk-level surrogate with $\rho_{t_0}^{\mathrm{corr}}$, we write a single training objective:
\begin{equation}
\mathcal{L}_{\mathrm{total}}
=
-J_{\mathrm{chunk}}^{\theta}
+\underbrace{
c_v\mathcal{L}_{\mathrm{value}}
-c_e \mathcal{H}\!\left[\pi_{\theta}^{\mathrm{co}}\right]
+\lambda_b \mathcal{L}_{\mathrm{bound}}}_{\substack{\text{Standard PPO Regularizers}}}
+\underbrace{\lambda_r \mathbb{E}_t\!\left[\|c_t\|_2^2\right]}_{\substack{\text{Corrector} \\ \text{Regularizer}}}.
\label{eq:acppo_corr_total_loss}
\end{equation}
The critic loss is
$\mathcal{L}_{\mathrm{value}}=\mathbb{E}_t[(V_\phi(s_t)-\hat R_t)^2]$,
$\mathcal{H}$ denotes the policy entropy,
$\mathcal{L}_{\mathrm{bound}}$ is the action-bound regularizer~\citep{makoviychuk2021isaac},
and the corrector regularizer term penalizes large corrections. In the planner pass, only $u_{t_0,k}$ receives actor gradients; in the corrector/value pass, $c_t$, $\sigma_t$, and $V_\phi$ are updated, using the two Adam updates shown in Algorithm~\ref{alg:acppo_corr}.

\paragraph{Value warm-up.}
In all experiments, we use a short warm-up stage before full joint training.
Specifically, we freeze the chunk planner for the first $N/20$ epochs and continue optimizing the corrector/value branch; after warm-up, we unfreeze the planner and train all modules jointly with separate Adam optimizers.
This allows the value function to create stable advantage estimates before the chunk-level planner begins to learn from it.

\section{Experiments}

\paragraph{Benchmark.} We evaluate ACPPO-Corr on 25 simulated robotics tasks, comprising 9 tasks from IsaacGym~\citep{makoviychuk2021isaac} and 16 tasks from Bi-DexHands~\citep{chen2022towards}. 
The benchmark spans locomotion, arm manipulation, and dexterous hand-object interaction with different horizons and contact dynamics. 
This diversity is particularly important in our setting, since the effect of action chunking and reduced decision frequency can vary across different tasks.

\paragraph{Algorithms.} We compare ACPPO-Corr against five baselines: PPO~\citep{schulman2017proximal}, the stepwise on-policy reference; PPO-Repeat~\citep{sharma2017learning}, PPO with repeated actions, which isolates the effect of reduced decision frequency; SAC~\citep{haarnoja2018soft}, a standard off-policy actor-critic; QC-FQL~\citep{li2025reinforcement}, the action chunking RL comparator; and ACPPO, which removes the proposed feedback corrector.
The chunk length and action repetition frequency are fixed to $h=4$.
In implementation, ACPPO-Corr uses a reduced-width planner/corrector design to keep model size comparable to PPO.
Appendix~\ref{app:model_size} reports the model sizes and the computational efficiency of each algorithm.

\begin{figure}[t]
    \centering
    \begin{subfigure}[b]{0.54\linewidth}
        \centering
        \parbox[c][4.5cm][c]{\linewidth}{%
            \centering
            \includegraphics[width=\linewidth]{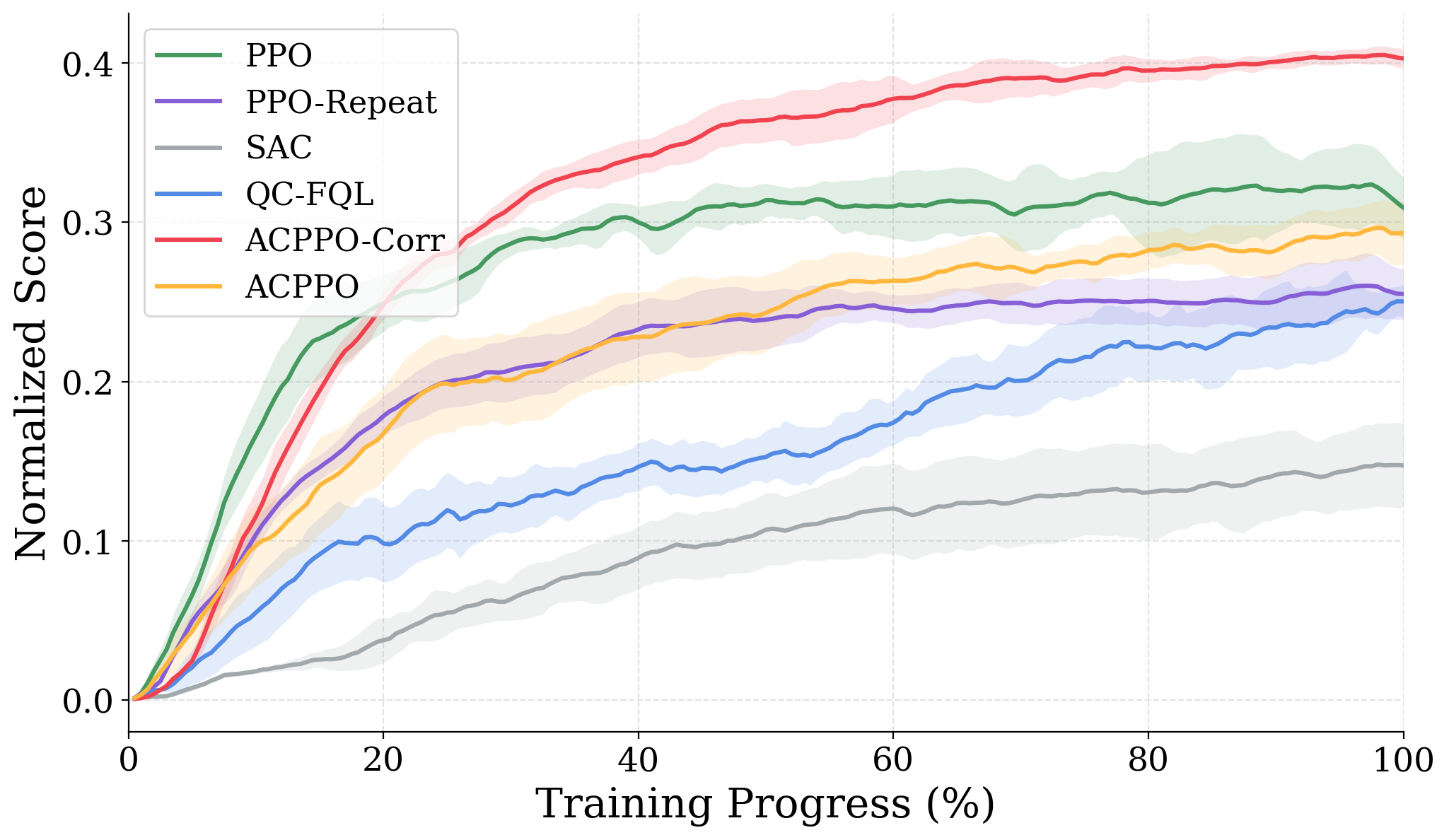}
        }
        \caption{Training Curve (25 Tasks)}
        \label{fig:benchmark_curve}
    \end{subfigure}
    \hfill
    \begin{subfigure}[b]{0.44\linewidth}
        \centering
        \parbox[c][4.5cm][c]{\linewidth}{%
            \centering
            \setlength{\tabcolsep}{4pt}
            \renewcommand{\arraystretch}{1.08}
            \footnotesize
            \begin{tabular}{@{}lcc@{}}
            \toprule
            \textbf{Method} & \textbf{Sensitive} & \textbf{Neutral} \\
            \midrule
            PPO         & 1.00 {\scriptsize [1.00, 1.00]} & 1.00 {\scriptsize [1.00, 1.00]} \\
            PPO-Repeat  & 0.61 {\scriptsize [0.55, 0.68]} & 0.97 {\scriptsize [0.92, 1.02]} \\
            SAC         & 0.33 {\scriptsize [0.21, 0.44]} & 0.59 {\scriptsize [0.49, 0.68]} \\
            QC-FQL      & 0.76 {\scriptsize [0.64, 0.87]} & 0.77 {\scriptsize [0.66, 0.87]} \\
            ACPPO       & 0.92 {\scriptsize [0.87, 0.97]} & 0.97 {\scriptsize [0.85, 1.10]} \\
            ACPPO-Corr  & \textbf{1.33 {\scriptsize [1.24, 1.43]}} & \textbf{1.24 {\scriptsize [1.15, 1.34]}} \\
            \bottomrule
            \end{tabular}
        }
        \caption{Frequency-Sensitivity Split Performance}
        \label{fig:benchmark_table}
    \end{subfigure}
    
    \caption{Benchmark results. \textbf{(a)} Normalized IQM Scores over training across all 25 tasks. Shaded regions show pointwise 95\% seed-bootstrap intervals with tasks fixed. ACPPO-Corr achieves best aggregated performance, improving over PPO by $30.4\%$ in final normalized IQM. \textbf{(b)} Final relative IQM with 95\% seed-bootstrap intervals with tasks fixed on the frequency-sensitive and frequency-neutral subsets, computed as the ratio to PPO's IQM on the same task set.}
    \label{fig:benchmark_results}
\end{figure}

\paragraph{Hyperparameters.}
For PPO-family baselines and SAC, we use the published task-specific hyperparameters from prior benchmark implementations~\cite{lee2024going, chen2022towards}. 
ACPPO-Corr inherits the PPO hyperparameters and introduces the corrector regularization weight $\lambda_r$. 
We select $\lambda_r \in \{0.03, 0.1, 0.3\}$ using preliminary per-task tuning runs, then fix the selected value for the reported evaluation. To match the tuning budget, we tune the QC-FQL behavior regularization coefficient $\alpha \in \{1,3,10\}$ using the same protocol.
Details are given in Appendix~\ref{app:hparams}, and full hyperparameters are given in Appendix~\ref{app:per_task_hyperparams}.

\paragraph{Metrics.}
Based on the task success indicator in IsaacGym and Bi-DexHands, we use two different metrics. 
For goal reaching tasks, where performance is defined by completion of the target behavior, we use the final success rate $(0-1)$. 
For progress-based control tasks, where performance is captured by the accumulated reward, we use episodic return. We then normalize each task score as
$
R_{\mathrm{norm}} = (R - R_{\mathrm{low}})/(R_{\mathrm{high}} - R_{\mathrm{low}}),
$
where $R_{\mathrm{low}}$ and $R_{\mathrm{high}}$ are the task-specific lower and upper bound values reported at Appendix~\ref{app:per_task_hyperparams}. 
Each method is evaluated with 5 random seeds on every task, and we report the interquartile mean (IQM)~\citep{agarwal2021deep} across all tasks.
For the final relative IQM, we compute the ratio between a method's final IQM and PPO's final IQM on the same task set.

\subsection{Benchmark Results}

\begin{figure}
    \centering
    \includegraphics[width=0.9\linewidth]{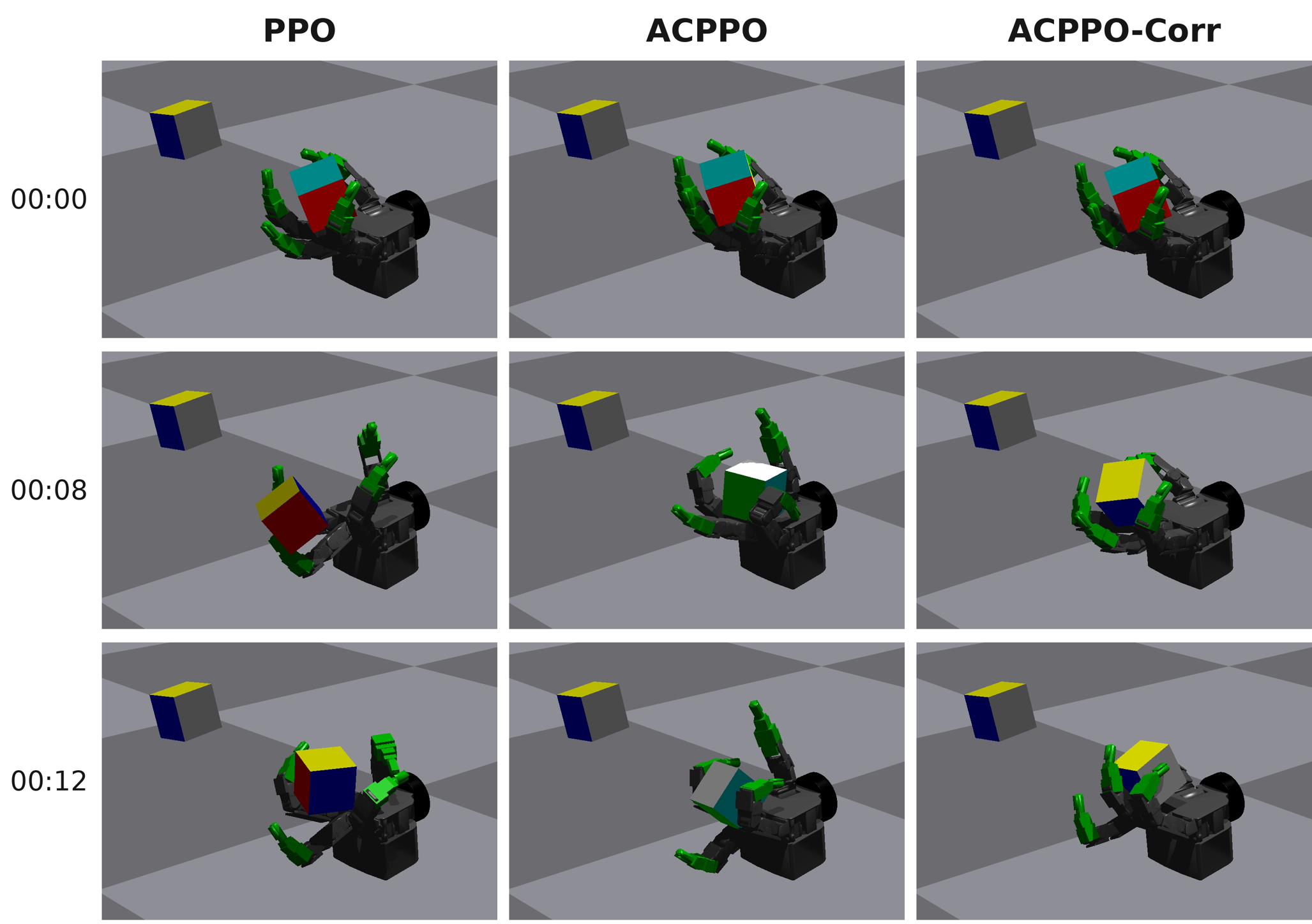}
    \caption{Rollout snapshots of PPO, ACPPO, and ACPPO-Corr on
    \texttt{AllegroHand}. ACPPO-Corr reaches the target orientation the fastest, where ACPPO drops the cube and PPO requires additional 128 environment interactions to reach the target.}
    \label{fig:visualization}
\end{figure}

\paragraph{Overall Performance.}
Figure~\ref{fig:benchmark_curve} summarizes the main benchmark results. ACPPO-Corr achieves best aggregated performance, improving over PPO by $30.4\%$ in final normalized IQM.
Both ACPPO and PPO-Repeat underperform PPO on the full benchmark, indicating that the gains of ACPPO-Corr cannot be explained by reduced decision frequency alone. 
ACPPO-Corr also substantially outperforms QC-FQL, our main chunked-action RL baseline, suggesting that actor-side chunking with a standard state-value critic is effective when paired with stepwise feedback correction in this online PPO setting.
Soft Actor-Critic~\citep{haarnoja2018soft} performs poorly overall, consistent with prior reports on Bi-DexHands~\citep{chen2022towards}; this reflects the difficulty of off-policy Q-learning in high-dimensional, contact-rich continuous action spaces.
The full learning curves are shown in Appendix~\ref{app:full_training_curve}, and the visualizations are displayed in Figure~\ref{fig:visualization}.

\paragraph{Frequency Sensitivity.}
Prior work~\citep{sharma2017learning, karimi2023dynamic} has shown that performance can depend strongly on interaction frequency and action repetition. 
Therefore, we partition tasks according to their sensitivity to reduced decision frequency. 
For each task $e$, we compute
\[
\Delta_e^{\mathrm{freq}} = \left|\frac{S_e^{\text{PPO}} - S_e^{\text{PPO-Repeat}}}{S_e^{\text{PPO}}}\right|,
\]
where $S_e$ denotes the final normalized score (without IQM) on task $e$. 
A larger $\Delta_e^{\mathrm{freq}}$ indicates that the task is more sensitive to changes in decision frequency. 
We exclude four tasks with near-zero PPO scores to ensure numerical stability of this ratio. 
Using a threshold of $\Delta_e^{\mathrm{freq}} > 0.3$, we split the remaining 21 tasks into 11 frequency-sensitive tasks and 10 frequency-neutral tasks.
The frequency-sensitivity split results are elaborated in Appendix~\ref{app:benchmarks}.

Figure~\ref{fig:benchmark_table} reports the final relative IQM under this split, computed as the ratio of IQMs within each subset. 
On frequency-neutral tasks, PPO and PPO-Repeat perform similarly, as expected when reduced decision frequency has limited effect.
On frequency-sensitive tasks, PPO-Repeat suffers a large performance drop, confirming that naive action repetition can be harmful when tasks require frequent feedback. 
Using the win/tie/loss defined in Appendix~\ref{app:benchmarks}, ACPPO-Corr is within the top group on 10/10 frequency-neutral tasks and 9/11 frequency-sensitive tasks.
This suggests that ACPPO-Corr preserves the benefits of chunk-level planning while avoiding the performance degradation caused by open-loop execution: the stepwise corrector provides within-chunk reactivity on tasks where reduced decision frequency is harmful.

\begin{figure}[t]
    \centering
    \begin{subfigure}[b]{0.48\linewidth}
        \centering
        \includegraphics[width=\linewidth]{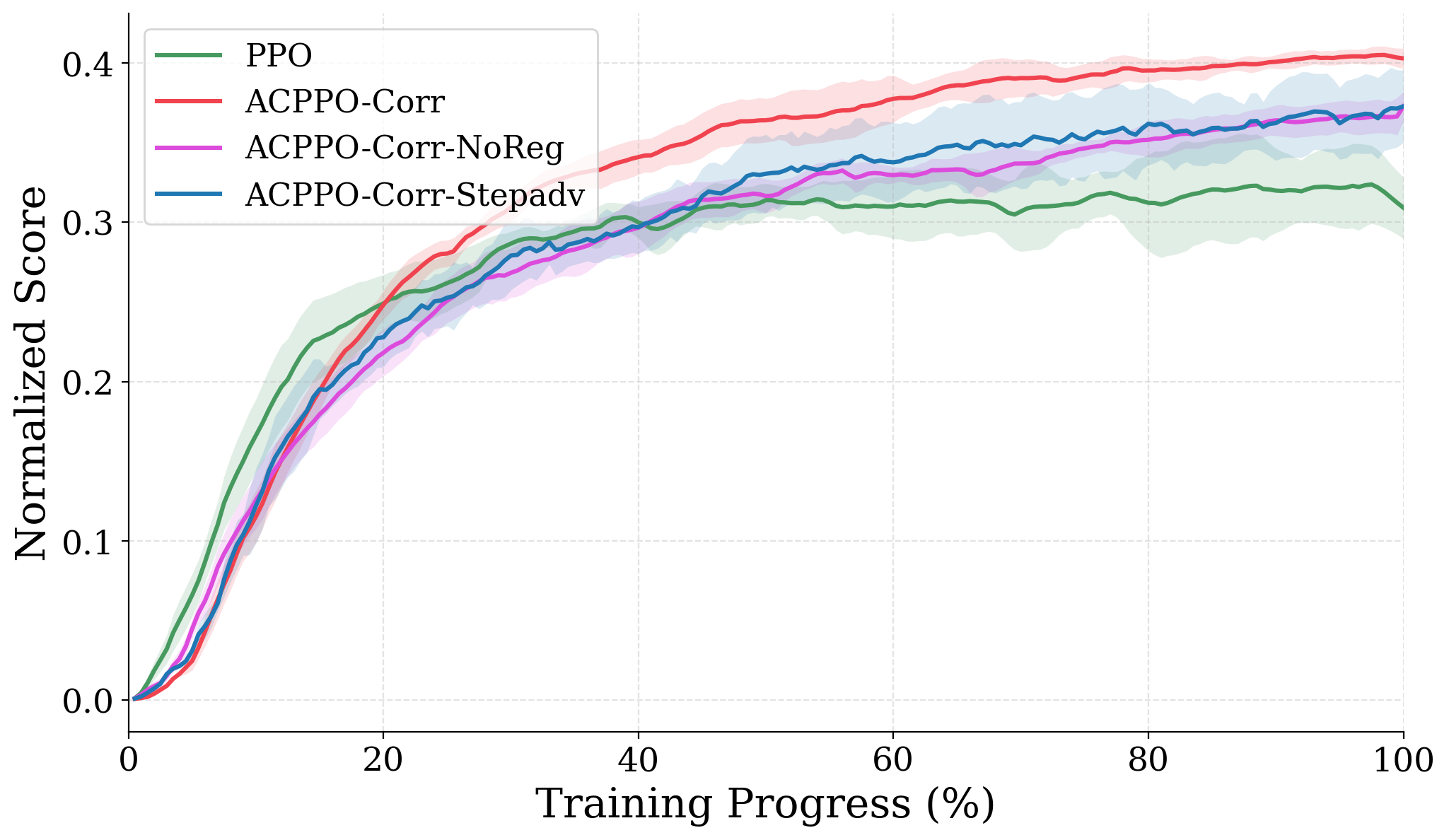}
        \caption{Advantage estimate and corrector regularization}
        \label{fig:noreg_ablation}
    \end{subfigure}
    \hfill
    \begin{subfigure}[b]{0.48\linewidth}
        \centering
        \includegraphics[width=\linewidth]{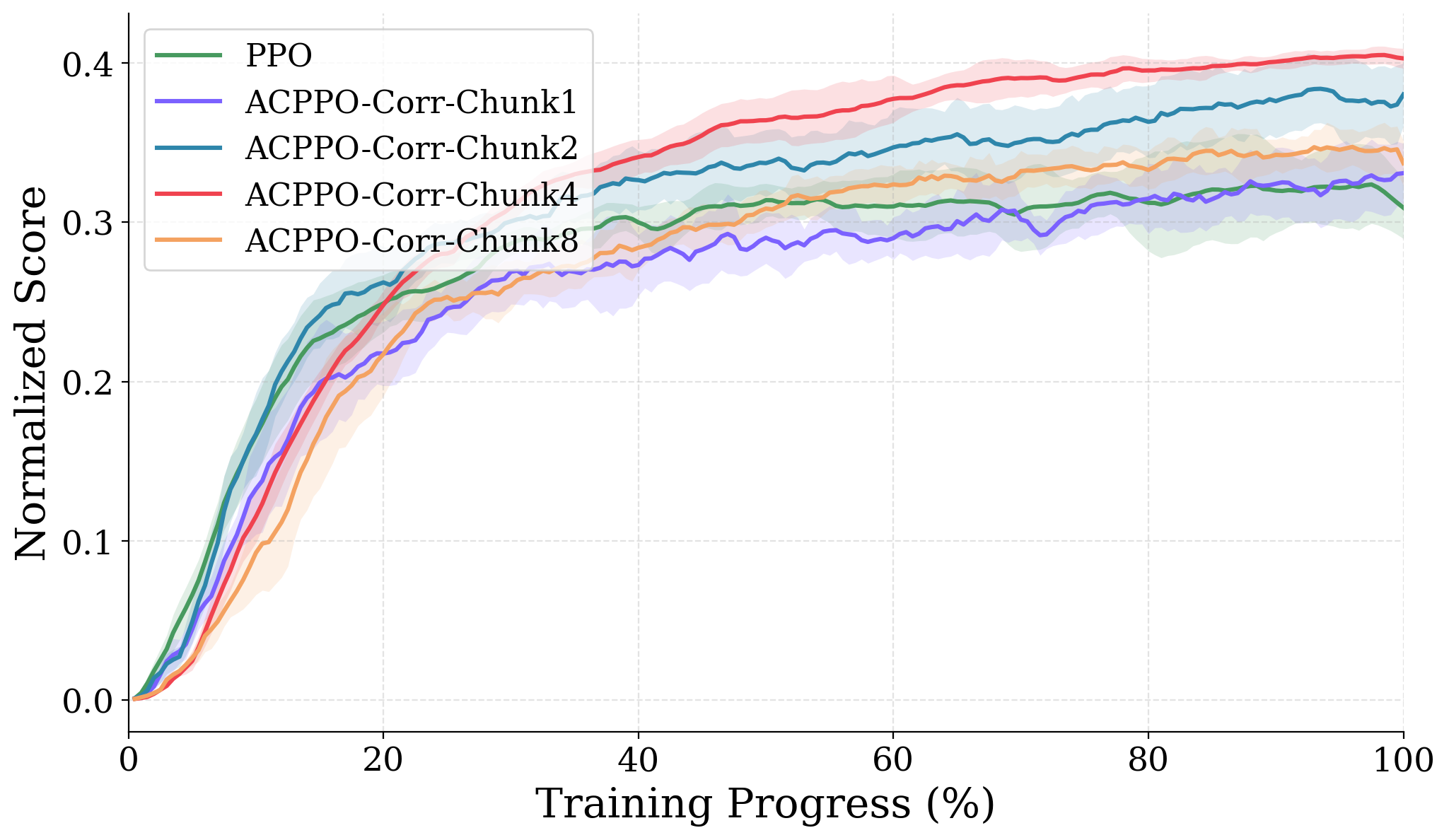}
        \caption{Sensitivity to chunk length}
        \label{fig:chunk_length_ablation}
    \end{subfigure}
    
    \caption{Ablations on chunked advantage, corrector regularization, and chunk length. (a) Chunked GAE and corrector regularization improve both performance and stability. (b) $h=1$ removes multi-step chunking while retaining the ACPPO-Corr pipeline. Moderate chunk lengths perform best, with $h=4$ giving the strongest aggregate performance.}
\end{figure}

\subsection{Ablation Study}
\label{sec:ablation}
\paragraph{Effect of the corrector regularizer.} 
The corrector regularizer in Eq.~\ref{eq:acppo_corr_total_loss} prevents the corrector from dominating the action plans.
Without this term, the corrector can absorb most of the action signal, causing the chunked planner to contribute little and making ACPPO-Corr behave like a mostly stepwise reactive policy.
ACPPO-Corr-NoReg of Figure~\ref{fig:noreg_ablation} compares ACPPO-Corr with and without the corrector regularizer.
The regularized variant achieves stronger aggregate performance, showing that balancing the planner and corrector is crucial for effective chunked control.

\paragraph{Effect of within-chunk feedback.}
To ablate the observations received during chunk
execution, we replace the corrector's current-state input $s_t$
with the chunk-start state $s_{t_0}$, keeping the architecture
unchanged. Applying this change only at evaluation retains
$84.0\%$ of ACPPO-Corr's performance, and training from scratch retains $88.1\%$ of ACPPO-Corr's performance. Retraining partially recovers performance, but the remaining gap supports the benefit of
within-chunk feedback.

\paragraph{Effect of the chunked advantage.}
We next ablate the chunked advantage used to train the ACPPO-Corr policy.
We compare ACPPO-Corr, which uses Eq.~\ref{eq:chunk_gae_impl} as the advantage estimate against a variant that replaces the chunked advantage with standard stepwise GAE.
As shown in Figure~\ref{fig:noreg_ablation}, ACPPO-Corr achieves stronger aggregate performance than ACPPO-Corr-Stepadv.
This suggests that the policy-gradient signal should match the temporal abstraction of the actor: because the planner selects an entire action chunk, assigning a single temporally extended advantage to the chunk decision provides more coherent credit assignment than using separate stepwise advantages. 
This result is also consistent with the analysis in Appendix~\ref{app:chunked_advantage_bias}, where the chunked estimator reduces reliance on intermediate value estimates, but uses a coarser and
potentially higher-variance chunk-level credit signal.

\paragraph{Sensitivity to the chunk length.}
To study the effect of the chunk length $h$, we evaluate chunk lengths $h \in \{1,2,4,8\}$ and compare it to the performance of PPO.
Since the rollout horizon length is 8 for most environments in the task set, we restrict the comparison to chunk lengths that divide this horizon.
The $h=1$ setting removes chunking while retaining the ACPPO-Corr pipeline: it analyzes whether the gains are explained by decomposing the policy into two branches.
Figure~\ref{fig:chunk_length_ablation} shows that $h=1$ improves only marginally over PPO and remains below the chunked variants. 
Performance improves substantially for $h>1$, with $h=4$ achieving the strongest aggregate performance. 
This suggests that the main gains of ACPPO-Corr come from combining chunk-level planning with stepwise feedback correction, rather than from the corrector alone.

\begin{figure}
    \centering
    \includegraphics[width=\linewidth]{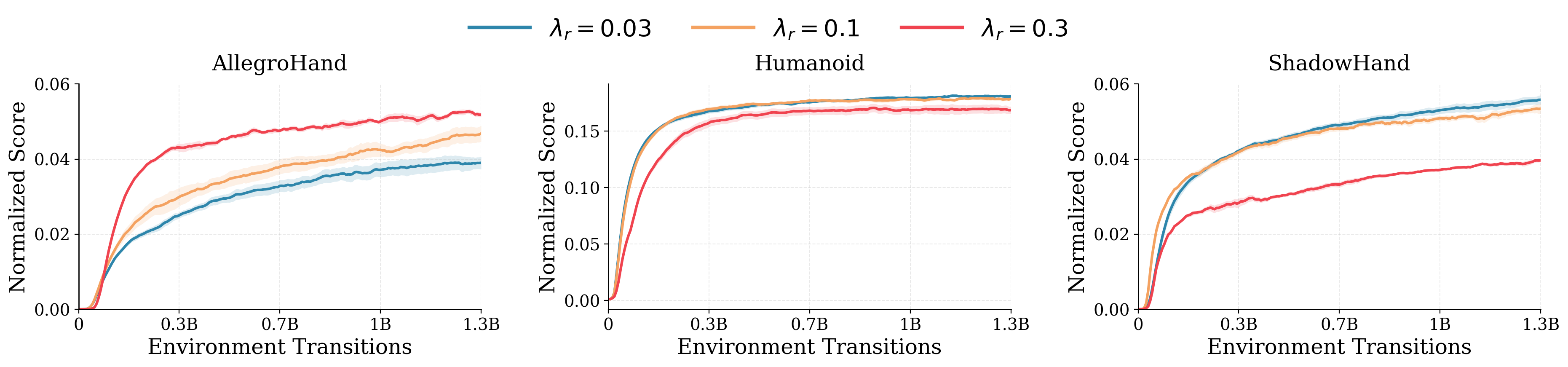}
    \caption{Sensitivity to the corrector regularizer $\lambda_r$ on three representative tasks. Different tasks prefer different regularization strengths: $\lambda_r$ is a relevant method-specific hyperparameter.}
    \label{fig:lambda_ablation}
\end{figure}

\paragraph{Sensitivity to the corrector regularizer hyperparameter.}
To study the effect of the corrector regularization coefficient $\lambda_r$, we select three representative tasks and sweep $\lambda_r \in \{0.03, 0.1, 0.3\}$. Figure~\ref{fig:lambda_ablation} shows that the best-performing value is task-dependent: stronger regularization works better on AllegroHand, while weaker regularization performs better on Humanoid and ShadowHand.
This sweep shows that the appropriate strength of the corrector regularizer varies across tasks.

To visualize how $\lambda_r$ changes the division between the planner and the corrector, we also report the final correction ratio
$
r_{\mathrm{corr}}
=
\|c_{t_0+k}\|_2/\|u_{t_0,k}\|_2.
$
Table~\ref{tab:corrector_ratio} shows that larger $\lambda_r$ consistently reduces $r_{\mathrm{corr}}$, confirming that the regularizer prevents the feedback corrector from dominating the chunk planner.

\begin{table}[t]
\centering

\begin{minipage}[t]{0.48\linewidth}
    \centering
    \caption{Final correction ratios across tasks and
    regularization strengths $\lambda_r$.}
    \label{tab:corrector_ratio}
    \small
    \vspace{3pt}
    \setlength{\tabcolsep}{3pt}
    \begin{tabular}{@{}lccc@{}}
        \toprule
        Task & $\lambda_r=0.03$ & $\lambda_r=0.1$
             & $\lambda_r=0.3$ \\
        \midrule
        AllegroHand & 0.222 & 0.053 & 0.007 \\
        Humanoid    & 0.021 & 0.009 & 0.002 \\
        ShadowHand  & 0.163 & 0.029 & 0.009 \\
        \bottomrule
    \end{tabular}
\end{minipage}\hfill
\begin{minipage}[t]{0.48\linewidth}
    \centering
    \caption{Longer-horizon evaluation.
    Entries are the means of per-task score ratios to PPO.}
    \label{tab:long_chunks}
    \small
    \vspace{3pt}
    \setlength{\tabcolsep}{3pt}
    \begin{tabular}{@{}lccc@{}}
        \toprule
        Method & $h=4$ & $h=8$ & $h=16$ \\
        \midrule
        ACPPO-Corr & 1.39 & 1.32 & 1.21 \\
        ACPPO      & 1.14 & 1.07 & 0.68 \\
        QC-FQL     & 0.88 & 0.87 & 0.29 \\
        \bottomrule
    \end{tabular}
\end{minipage}
\end{table}

\paragraph{Longer chunk horizons.}
We extend the chunk-length evaluation to $h\in\{4,8,16\}$ on the five tasks whose rollout horizons support these lengths.
As shown in Table~\ref{tab:long_chunks}, ACPPO-Corr degrades less as $h$ increases, retaining a mean score ratio to PPO of $1.21$ at $h=16$, compared with $0.68$ for ACPPO and $0.29$ for QC-FQL.
We additionally evaluate $h=32$ on FrankaCubeStack and Humanoid in Table~\ref{tab:long_chunks_full}.
ACPPO-Corr achieves nontrivial scores, while ACPPO and QC-FQL each obtain at most $0.02$ on either task.
These results support the benefit of feedback correction over longer execution windows, although moderate chunk lengths remain preferable.

\section{Conclusion \& Limitations}
\label{sec:Conclusion}

We presented ACPPO, an action-chunking variant of PPO with a standard state-value critic, and ACPPO-Corr, which augments chunk-level planning with a stepwise feedback corrector. 
This design brings temporal abstraction into a fully online, on-policy PPO framework while providing within-chunk reactivity which is important in contact-rich robotic control. 
Across 25 simulated robotics tasks spanning IsaacGym and Bi-DexHands, ACPPO-Corr achieved the strongest overall benchmark performance and remained best on both frequency-sensitive and frequency-neutral task subsets, suggesting the impact of action chunking and closed-loop action corrections. 
The open-loop ACPPO baseline underperforms PPO on the full benchmark, indicating that actor-side chunking alone is not sufficient for these contact-rich tasks.
Our ablations further indicate that moderate chunk lengths are most effective and that regularizing the corrector helps balance the division between the chunk planner and the feedback corrector. 
Overall, these results suggest that action chunking can be made effective in PPO when open-loop temporal structure is paired with closed-loop correction. 

However, several limitations remain. First, our study is limited to continuous robotics control tasks: ACPPO-Corr assumes a continuous action space where additive corrections and magnitude penalties are meaningful, and therefore does not directly apply to discrete or semantically discontinuous action spaces such as token-level NLP. 
Furthermore, our experiments are conducted only in simulated robotics environments with state-based observations. 
Finally, the chunked importance sampling ratio may become less stable as the chunk length increases. 
While clipping remained stable for short chunk lengths, increased clipping was observed in $h=8$. 
Applying ACPPO-Corr to substantially longer chunks may require additional stabilization.
Important directions for future work include adaptive chunk lengths, sim-to-real transfer, and extensions to latent-action formulations for application beyond continuous control.

\bibliographystyle{unsrtnat}
\bibliography{neurips_2026}

\newpage
\appendix

\section{Experiment Details}
\subsection{Benchmarks}
\label{app:benchmarks}

We use the composition of 9 tasks from IsaacGym~\citep{makoviychuk2021isaac}, and 16 tasks from Bi-DexHands~\citep{chen2022towards}.
For the frequency-sensitivity analysis only, we exclude four tasks whose PPO scores are near zero, since the relative frequency-sensitivity ratio $\Delta_e^{\mathrm{freq}}$ becomes numerically unstable. 
The total task sets and the frequency-sensitivity splits are displayed in Table~\ref{tab:task_split_wtl}.

We also report a per-task win/tie/loss status for ACPPO-Corr. 
For each task $e$, let $S_{e,m}$ denote the final normalized score of method $m$, averaged over five seeds.
We compare ACPPO-Corr against the strongest baseline method on the same task:
\[
\Delta^{\mathrm{WTL}}_e =
S_{e,\mathrm{ACPPO\text{-}Corr}}
-
\max_{m \neq \mathrm{ACPPO\text{-}Corr}} S_{e,m}.
\]
We mark a task as a win if $\Delta^{\mathrm{WTL}}_e > 0.03$, a tie if
$|\Delta^{\mathrm{WTL}}_e| \leq 0.03$, and a loss if
$\Delta^{\mathrm{WTL}}_e < -0.03$.

\begin{table}[H]
\centering
\small
\setlength{\tabcolsep}{3pt}
\caption{
Task assignment for the frequency-sensitivity split and per-task win/tie/loss
status of ACPPO-Corr. Superscripts denote ACPPO-Corr's status against the
strongest competing method on each task: W = win, T = tie, L = loss.
}
\label{tab:task_split_wtl}
\begin{tabular}{p{0.13\linewidth}p{0.68\linewidth}cc}
\toprule
Group & Tasks & \# & W/T/L \\
\midrule
Neutral
&
FrankaCabinet$^{W}$,
Ingenuity$^{T}$,
ShadowHandDoorOpenInward$^{W}$,
ShadowHandDoorCloseInward$^{T}$,
ShadowHandDoorOpenOutward$^{T}$,
ShadowHandScissors$^{W}$,
QuadCopter$^{T}$,
ShadowHandGraspAndPlace$^{W}$,
ShadowHandOver$^{T}$,
ShadowHandBlockStack$^{W}$
& 10 & 5/5/0 \\
\midrule
Sensitive
&
ShadowHandSpin$^{W}$,
ShadowHandBottleCap$^{L}$,
Humanoid$^{W}$,
Anymal$^{T}$,
ShadowHand$^{W}$,
AllegroHand$^{W}$,
Ant$^{W}$,
ShadowHandUpsideDown$^{L}$,
ShadowHandCatchOver2Underarm$^{W}$,
ShadowHandCatchAbreast$^{W}$,
FrankaCubeStack$^{T}$
& 11 & 7/2/2 \\
\midrule
Excluded
&
ShadowHandLiftUnderarm$^{W}$,
ShadowHandDoorCloseOutward$^{L}$,
ShadowHandKettle$^{L}$,
ShadowHandPushBlock$^{T}$
& 4 & 1/1/2 \\
\midrule
All tasks & -- & 25 & 13/8/4 \\
\bottomrule
\end{tabular}
\end{table}

\subsection{Model Size and Computational Overhead}
\label{app:model_size}

The MLP widths are task-dependent and follow the standard configurations used in~\citet{lee2024going}. 
For PPO and PPO-Repeat, we inherit the default task-specific PPO actor and state-value critic architectures. 
ACPPO keeps the same PPO architecture, except that the actor output dimension is expanded from $d_a$ to $h d_a$ to predict an action chunk. 

For methods with multiple actor or critic branches, we use reduced-width MLPs, with the first layer of the MLP divided in half to keep the total model size comparable. 
Table~\ref{tab:param_accounting} summarizes the network components used by each method. 
Table~\ref{tab:efficiency} reports representative parameter counts and training throughput under our benchmark implementation.
All experiments were conducted on a single RTX A6000 GPU.

\begin{table}[H]
    \centering
    \caption{Network components used per algorithm. 
    Halved MLP configuration is used for methods with multiple actor or critic branches.}
    \label{tab:param_accounting}
    \small
    \begin{tabular}{ll}
        \toprule
        \textbf{Algorithm} & \textbf{Network components} \\
        \midrule
        PPO & 1 full-size actor + 1 full-size value critic \\
        ACPPO & 1 full-size chunked actor + 1 full-size value critic \\
        ACPPO-Corr & 1 chunk planner + 1 corrector + 1 full-size value critic \\
        SAC & 1 actor + 2 Q-functions + 2 target Q-functions \\
        QC-FQL & 1 chunked actor + 2 Q-functions + 2 target Q-functions + 1 flow network \\
        \bottomrule
    \end{tabular}
\end{table}

\begin{table}[H]
    \centering
    \caption{Training throughput and model size. 
    Steps/sec includes rollout and optimization. 
    Parameter counts include policy/value networks and, for off-policy methods, target critic networks used during training.}
    \label{tab:efficiency}
    \small
    \begin{tabular}{lrr}
        \toprule
        \textbf{Algorithm} & \textbf{Steps/sec} & \textbf{Parameters} \\
        \midrule
        PPO        & 47,894 & 1,463,785 \\
        ACPPO      & 45,579 & 1,484,065 \\
        ACPPO-Corr & 41,714 & 1,557,258 \\
        SAC        & 39,049 & 2,024,428 \\
        QC-FQL     & 34,595 & 3,287,492 \\
        \bottomrule
    \end{tabular}
\end{table}

\subsection{Hyperparameter Selection}
\label{app:hparams}

Unless otherwise stated, all chunked methods use chunk length $h=4$ in the main benchmark. 
The effect of varying $h$ is studied separately in Section~\ref{sec:ablation}.

For PPO and SAC, we inherit the task-specific hyperparameters from~\citet{lee2024going} and~\citet{chen2022towards}. 
PPO-Repeat and ACPPO use the same hyperparameters and network architecture as PPO.

ACPPO-Corr introduces one additional hyperparameter, the corrector regularization coefficient $\lambda_r$. For each environment, we select $\lambda_r$ from $ \{0.03, 0.1, 0.3\} $ using preliminary tuning runs, then use the selected value for the reported five seed runs.

For QC-FQL, we use the same three-value tuning budget and select the behavior regularization coefficient from $
\alpha \in \{1, 3, 10\}
$
following the same procedure as ACPPO-Corr.
We find that $\alpha=1$ performs best in most environments. 
This is expected in our fully online setting: early rollouts are noisy and non-expert, so strong behavior cloning to the replay buffer can constrain policy improvement. 
Reducing the behavior regularization allows the policy to deviate from early low-quality behavior while still retaining the stabilizing effect of the regularizer.
The selected coefficient for each environment was chosen using preliminary tuning runs and then fixed for the reported five-seed evaluation.
The selected values per environment for both algorithms are reported in Appendix~\ref{app:per_task_hyperparams}.

\subsection{Action Chunking RL Baselines}

\label{app:baseline_discussion}
Several related chunk-correction and real-time execution methods are not included as direct baselines because they assume a different training interface from our benchmark. 
In particular, many methods operate on frozen behavior-cloned chunked policies, require demonstrations or pretrained VLA/diffusion policies, use visual or tactile observations not available in our state-based benchmark, or modify test-time chunk resampling rather than online RL training from scratch. Including such methods would require adding offline data, pretrained policies, or different observation modalities, which would break the online state-based comparison. 
We therefore focus the main benchmark on methods that can be trained from scratch under a matched online protocol, and include QC-FQL as the closest chunked-action RL comparator.

\subsection{Aggregation}

For each method, task, and seed, we first normalize the evaluation curve using the task-specific bounds in Appendix~\ref{app:per_task_hyperparams}, smooth the curve, and interpolate it onto a common training-progress grid. 
If multiple runs share the same method-task-seed tuple, we average them before aggregation. 
Let $x_{m,e,s}(p)$ denote the resulting normalized score for method $m$, task $e$, seed $s$, and training progress $p$.

At each progress point $p$, we aggregate performance using the interquartile mean (IQM)~\citep{agarwal2021deep} over the set of task-seed scores
\[
\mathcal X_m(p)
=
\{x_{m,e,s}(p): e \in \mathcal E,\; s \in \mathcal S\}.
\]
The solid learning curve reports
\[
\mathrm{IQM}_m(p)
=
\mathrm{IQM}(\mathcal X_m(p)),
\]
where IQM is the mean of the middle 50\% of scores after sorting.

Uncertainty intervals are computed by seed bootstrap with the task set fixed. 
For each bootstrap replicate, we resample the five seed slots with replacement, recompute $\mathrm{IQM}_m(p)$ at every progress point using the resampled task-seed scores, and report the pointwise 2.5 and 97.5 percentiles. 
Thus, the shaded regions reflect variability due to random seeds under the fixed benchmark suite, not uncertainty over a broader distribution of tasks.

Final performance is computed analogously from the last finite value of each normalized task-seed curve. 
For PPO-normalized performance on a task subset $\mathcal E'$, we recompute the method IQM and PPO IQM under the same bootstrap seed resample and report
\[
\mathrm{RelIQM}_m
=
\frac{\mathrm{IQM}_m}{\mathrm{IQM}_{\mathrm{PPO}}}.
\]
Bootstrap intervals for relative IQM are obtained from the corresponding bootstrap distribution of this ratio.

\subsection{Full Training Curve}
\label{app:full_training_curve}

\begin{figure}[h]
    \centering
    \includegraphics[width=\linewidth]{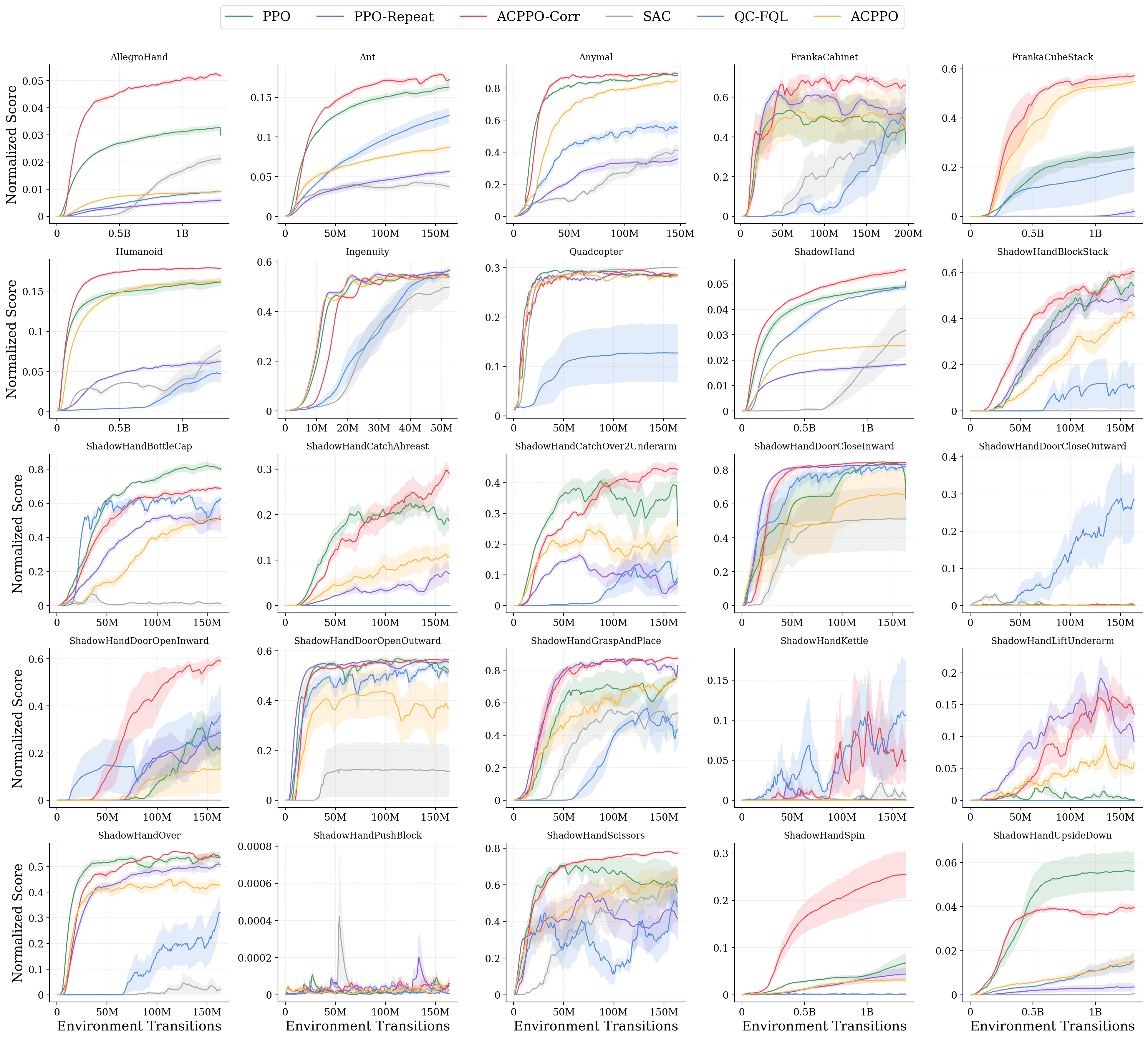}
    \caption{Full training curves on IsaacGym and Bi-DexHands.}
    \label{fig:fullexp}
\end{figure}

\subsection{Per Task Hyperparameters}
\label{app:per_task_hyperparams}

\begin{table}[H]
    \centering
    \caption{Per-task selected hyperparameters and normalization bounds. 
    $\lambda_r$ denotes the corrector regularization coefficient for ACPPO-Corr at each chunk length,
    $\alpha$ denotes the behavior regularization coefficient for QC-FQL, and 
    $R_{\mathrm{low}}, R_{\mathrm{high}}$ are the task-specific bounds used in normalized IQM computation.}
    \label{tab:task_hparams_norm}
    \scriptsize
    \setlength{\tabcolsep}{3.5pt}
    \renewcommand{\arraystretch}{1.08}
    \resizebox{\linewidth}{!}{
    \begin{tabular}{@{}lccccccc@{}}
        \toprule
        \textbf{Task} 
        & $\lambda_r$ $(h=1)$ 
        & $\lambda_r$ $(h=2)$ 
        & $\lambda_r$ $(h=4)$ 
        & $\lambda_r$ $(h=8)$ 
        & QC-FQL $\alpha$ 
        & $R_{\mathrm{low}}$ 
        & $R_{\mathrm{high}}$ \\
\midrule
        \texttt{FrankaCabinet}                  & 0.1  & 0.1  & 0.1  & 0.1  & 3  & 0  & 1 \\
        \texttt{ShadowHandDoorOpenInward}       & 0.1  & 0.3  & 0.1  & 0.1  & 1  & 0  & 1 \\
        \texttt{ShadowHandDoorCloseInward}      & 0.3  & 0.1  & 0.1  & 0.1  & 1  & 0  & 1 \\
        \texttt{ShadowHandDoorOpenOutward}      & 0.3  & 0.3  & 0.3  & 0.3  & 1  & 0  & 1 \\
        \texttt{ShadowHandScissors}             & 0.3  & 0.3  & 0.3  & 0.3  & 1  & 0  & 1 \\
        \texttt{Ingenuity}                      & 0.1  & 0.1  & 0.1  & 0.1  & 1  & 0  & 14000 \\
        \texttt{QuadCopter}                     & 0.3  & 0.3  & 0.3  & 0.3  & 1  & 0  & 1500 \\
        \texttt{ShadowHandGraspAndPlace}        & 0.1  & 0.1  & 0.1  & 0.1  & 1  & 0  & 1 \\
        \texttt{ShadowHandOver}                 & 0.1  & 0.1  & 0.1  & 0.03 & 1  & 0  & 1 \\
        \texttt{ShadowHandBlockStack}           & 0.3  & 0.3  & 0.3  & 0.1  & 1  & 0  & 1 \\
        \midrule
        \texttt{ShadowHandSpin}                 & 0.3  & 0.3  & 0.3  & 0.3  & 1  & 0  & 1 \\
        \texttt{ShadowHandBottleCap}            & 0.03 & 0.03 & 0.03 & 0.03 & 1  & 0  & 1 \\
        \texttt{Anymal}                         & 0.3  & 0.3  & 0.3  & 0.3  & 1  & 0  & 75 \\
        \texttt{ShadowHand}                     & 0.03 & 0.03 & 0.03 & 0.03 & 1  & 0  & 1 \\
        \texttt{Humanoid}                       & 0.1  & 0.1  & 0.1  & 0.1  & 10 & -1 & 62841 \\
        \texttt{Ant}                            & 0.3  & 0.3  & 0.3  & 0.3  & 1  & -2 & 61341 \\
        \texttt{ShadowHandUpsideDown}           & 0.1 & 0.03 & 0.03 & 0.03 & 1  & 0  & 1 \\
        \texttt{ShadowHandCatchOver2Underarm}   & 0.3  & 0.3  & 0.3  & 0.3  & 1  & 0  & 1 \\
        \texttt{AllegroHand}                    & 0.1  & 0.1  & 0.3  & 0.1  & 1  & 0  & 1 \\
        \texttt{FrankaCubeStack}                & 0.3  & 0.3  & 0.3  & 0.3  & 1  & 0  & 1 \\
        \texttt{ShadowHandCatchAbreast}         & 0.03 & 0.03 & 0.03 & 0.03 & 1  & 0  & 1 \\
        \midrule
        \texttt{ShadowHandLiftUnderarm}         & 0.1  & 0.1  & 0.3  & 0.3  & 1  & 0  & 1 \\
        \texttt{ShadowHandKettle}               & 0.3  & 0.3  & 0.3  & 0.3  & 1  & 0  & 1 \\
        \texttt{ShadowHandPushBlock}            & 0.3  & 0.3  & 0.3  & 0.3  & 1  & 0  & 1 \\
        \texttt{ShadowHandDoorCloseOutward}     & 0.3  & 0.3  & 0.3  & 0.3  & 1  & 0  & 1 \\
        \bottomrule
    \end{tabular}
    }
\end{table}

\subsection{Simulation and Control Frequencies}
\label{app:control_frequencies}

\begin{table}[ht]
\centering
\caption{Simulation and control timescales. The ShadowHand row includes all ShadowHand-prefixed tasks.}
\label{tab:control_frequencies}
\begin{tabular}{lrrrr}
\toprule
Task & Physics (Hz) & Action repeat & Control (Hz) & $h=4$ (s)\\
\midrule
AllegroHand & 120 & 4 & 30 & 0.133\\
Ant & 120 & 2 & 60 & 0.067\\
Anymal & 100 & 2 & 50 & 0.080\\
FrankaCabinet & 60 & 1 & 60 & 0.067\\
FrankaCubeStack & 120 & 2 & 60 & 0.067\\
Humanoid & 120 & 2 & 60 & 0.067\\
Ingenuity & 200 & 2 & 100 & 0.040\\
QuadCopter & 200 & 2 & 100 & 0.040\\
ShadowHand tasks & 120 & 2 & 60 & 0.067\\
\bottomrule
\end{tabular}
\end{table}

\subsection{Longer-horizon evaluation.}
Ingenuity, FrankaCabinet, and Ant use rollout horizon $T=16$; FrankaCubeStack and Humanoid use $T=32$.
All five tasks are evaluated at $h\in\{4,8,16\}$, and the latter two also support $h=32$.
Table~\ref{tab:long_chunks_full} gives the per-task score ratios to PPO.

\begin{table}[ht]
\centering
\caption{Per-task score ratios to PPO at longer chunk lengths. A dash denotes an unevaluated setting due to the rollout horizon.}
\label{tab:long_chunks_full}
\begin{tabular}{llrrrr}
\toprule
Method & Task & $h=4$ & $h=8$ & $h=16$ & $h=32$\\
\midrule
ACPPO-Corr & FrankaCubeStack & 2.21 & 2.17 & 1.72 & 1.09\\
 & Humanoid & 1.11 & 1.07 & 0.97 & 0.65\\
 & Ingenuity & 0.99 & 1.00 & 0.99 & --\\
 & FrankaCabinet & 1.56 & 1.39 & 1.48 & --\\
 & Ant & 1.06 & 0.97 & 0.88 & --\\
\midrule
ACPPO & FrankaCubeStack & 2.11 & 1.90 & 1.37 & 0.00\\
 & Humanoid & 1.00 & 0.71 & 0.52 & 0.02\\
 & Ingenuity & 0.99 & 0.57 & 0.37 & --\\
 & FrankaCabinet & 1.08 & 1.48 & 0.72 & --\\
 & Ant & 0.53 & 0.68 & 0.43 & --\\
\midrule
QC-FQL & FrankaCubeStack & 0.75 & 1.59 & 1.12 & 0.00\\
 & Humanoid & 0.29 & 0.53 & 0.20 & 0.01\\
 & Ingenuity & 1.06 & 0.78 & 0.01 & --\\
 & FrankaCabinet & 1.53 & 0.98 & 0.00 & --\\
 & Ant & 0.78 & 0.48 & 0.10 & --\\
\bottomrule
\end{tabular}
\end{table}

\subsection{Task dependence and failures.}
The benefit of ACPPO-Corr varies with the need for temporally
coherent actions and within-chunk feedback.
On \texttt{AllegroHand} and \texttt{ShadowHandSpin}, ACPPO-Corr achieves $1.45$
and $3.76$ times PPO's score, respectively.
Both tasks require coherent multi-step in-hand manipulation while continuously adapting to object slip and changing finger contacts.

On \texttt{FrankaCabinet} and \texttt{FrankaCubeStack}, open-loop ACPPO already
performs well: their lower-dimensional manipulation and relatively stable parallel-gripper contacts reduce the need for within-chunk correction.

ACPPO-Corr underperforms PPO on \texttt{ShadowHandUpsideDown} and
\texttt{ShadowHandBottleCap}, achieving $0.706$ and $0.829$ times PPO's score.
We hypothesize that failures in these tasks require abandoning the current manipulation plan. A slipping pen may require an immediate reconfiguration of all fingers, and a failed cap grasp may require releasing and establishing a new contact pose. However, since ACPPO-Corr generates actions by correcting the planned sequence, it may remain biased toward the current manipulation mode and therefore underperform PPO.

For \texttt{ShadowHandPushBlock}, all evaluated methods obtained zero success, consistent with the behavior observed in the original benchmark~\citep{lee2024going}. We suspect that the existing reward signal and exploration budget provide insufficient guidance for discovering the coordinated two-hand pushing behavior. However, because none of the evaluated methods solves the task, we cannot conclusively attribute the failure to a particular component of the algorithms.

For \texttt{ShadowHandDoorCloseOutward}, the case is different: QC-FQL shows nonzero performance, whereas the other evaluated methods remain near zero. One possibility is that QC-FQL’s flow-based actor better represents useful chunk distributions, or that its off-policy replay retains rare successful trajectories more effectively. However, QC-FQL differs in its policy class, replay-based data usage, Q-function optimization, and objective, making it difficult to explain it as a single factor.

\section{Closed-loop Chunked Importance Ratio}
\label{app:closed_loop_ratio}

\paragraph{Closed-loop chunk likelihood.}
Although the planner is evaluated only at chunk boundaries, ACPPO-Corr defines a valid stochastic policy at every control step. For a chunk starting at $t_0$, let $k=t-t_0$ and let $\mathcal{T}_{t_0}\subseteq\{0,\ldots,h-1\}$ denote the valid set of executed offsets before any mid-chunk termination. The planner outputs the chunk mean sequence
\[
\mathbf{u}_{\theta,t_0}=\bigl[u_{\theta,t_0,0},\ldots,u_{\theta,t_0,h-1}\bigr]
=\pi^{\mathrm{pl}}_{\theta}(s_{t_0}),
\]
and the corrector outputs a state-dependent correction and exploration scale
\[
(c_{\theta}(s_t),\sigma_{\theta}(s_t))=\pi^{\mathrm{co}}_{\theta}(s_t).
\]
The executed policy inside the chunk is therefore
\begin{equation*}
\pi_\theta(a_t \mid s_t,s_{t_0},k)
=
\mathcal{N}\!\left(
 a_t;
 u_{\theta,t_0,k}+c_{\theta}(s_t),
 \operatorname{diag}(\sigma_{\theta}(s_t)^2)
\right).
\end{equation*}
For a realized chunk prefix, the probability density of the sampled segment under $\pi_\theta$ is
\begin{equation}
p_\theta(\tau_{t_0}\mid s_{t_0})
=
\prod_{k\in\mathcal{T}_{t_0}}
\pi_\theta(a_{t_0+k}\mid s_{t_0+k},s_{t_0},k)
P(s_{t_0+k+1}\mid s_{t_0+k},a_{t_0+k}),
\label{eq:app_corr_traj_prob}
\end{equation}
where the transition kernel $P$ is independent of the policy parameters. Hence, for the same sampled trajectory segment, the likelihood ratio between the current policy and the behavior policy is
\begin{align*}
\frac{p_\theta(\tau_{t_0}\mid s_{t_0})}
     {p_{\theta_{\mathrm{old}}}(\tau_{t_0}\mid s_{t_0})}
&=
\prod_{k\in\mathcal{T}_{t_0}}
\frac{
\pi_\theta(a_{t_0+k}\mid s_{t_0+k},s_{t_0},k)
}{
\pi_{\theta_{\mathrm{old}}}(a_{t_0+k}\mid s_{t_0+k},s_{t_0},k)
}.
\end{align*}
Therefore, the product ratio in Eq.~\ref{eq:acppo_corr_ratio} is the correct likelihood ratio for the sampled closed-loop chunk, even though the corrector observes the current state inside the chunk. If the episode terminates inside a chunk, the product is taken only over the executed prefix $\mathcal{T}_{t_0}$ and the remaining offsets are masked.

\paragraph{Unclipped on-policy interpretation.}
For a chunk-start state $s_{t_0}$, define the $h$-step bootstrapped target
\begin{equation*}
G^{(h)}_{t_0}
=
\sum_{k=0}^{h-1}\gamma^k r_{t_0+k}
+
\gamma^h V^\pi(s_{t_0+h}),
\end{equation*}
and the corresponding chunk-level advantage
\begin{equation*}
A_h^\pi(s_{t_0},\tau_{t_0})
=
G^{(h)}_{t_0}-V^\pi(s_{t_0}).
\end{equation*}
Since the transition terms in Eq.~\ref{eq:app_corr_traj_prob} do not depend on $\theta$, the score of the sampled chunk prefix is
\begin{equation*}
\nabla_\theta \log p_\theta(\tau_{t_0}\mid s_{t_0})
=
\sum_{k\in\mathcal{T}_{t_0}}
\nabla_\theta
\log \pi_\theta(a_{t_0+k}\mid s_{t_0+k},s_{t_0},k).
\end{equation*}
Therefore, in the unclipped on-policy actor update, and treating the value target as fixed as in standard actor--critic methods, the chunk-level likelihood-ratio contribution takes the form
\begin{equation*}
\left(
\sum_{k\in\mathcal{T}_{t_0}}
\nabla_\theta
\log \pi_\theta(a_{t_0+k}\mid s_{t_0+k},s_{t_0},k)
\right)
A_h^\pi(s_{t_0},\tau_{t_0}).
\end{equation*}

\section{Chunked Advantage: Value-Error Cancellation and Tradeoff}
\label{app:chunked_advantage_bias}

\paragraph{Setup.}
Let $V_*^\pi$ denote the projected state value from Section~\ref{sec:projected_state_critic}. For notational simplicity,
we write it as $V^\pi$ in this appendix, and define the critic approximation error
$e(s):=V_\phi(s)-V^\pi(s)$.
Let
\begin{equation*}
\delta_t := r_t + \gamma V_\phi(s_{t+1}) - V_\phi(s_t),
\qquad
\tilde\delta_t := r_t + \gamma V^\pi(s_{t+1}) - V^\pi(s_t).
\end{equation*}
Then
\begin{equation}
\delta_t-\tilde\delta_t
=
\gamma e(s_{t+1})-e(s_t).
\label{eq:app_td_error_difference}
\end{equation}

\paragraph{Stepwise GAE.}
Let $\alpha:=\gamma\lambda$. Standard stepwise GAE and its oracle counterpart are
\begin{equation*}
\hat A_t^{\mathrm{GAE}}
:=
\sum_{j=0}^{\infty}\alpha^j\delta_{t+j},
\qquad
\tilde A_t^{\mathrm{GAE}}
:=
\sum_{j=0}^{\infty}\alpha^j\tilde\delta_{t+j}.
\end{equation*}
Using Eq.~\ref{eq:app_td_error_difference}, their difference is
\begin{align}
\hat A_t^{\mathrm{GAE}}-\tilde A_t^{\mathrm{GAE}}
&=
\sum_{j=0}^{\infty}\alpha^j
\left(\gamma e(s_{t+j+1})-e(s_{t+j})\right) \notag\\
&=
-e(s_t)
+
\gamma(1-\lambda)
\sum_{j=1}^{\infty}
(\gamma\lambda)^{j-1}e(s_{t+j}).
\label{eq:app_step_gae_error}
\end{align}
The first term is the start-state value error. 
The second term is the recursive bootstrap error from future critic errors.
Define
\begin{equation*}
F_t
:=
\gamma(1-\lambda)
\sum_{j=1}^{\infty}
(\gamma\lambda)^{j-1}e(s_{t+j}).
\end{equation*}
Then
\begin{equation*}
\hat A_t^{\mathrm{GAE}}-\tilde A_t^{\mathrm{GAE}}
=
-e(s_t)+F_t.
\end{equation*}
If $\|e\|_\infty\leq\epsilon$, the recursive term is bounded by
\begin{equation}
|F_t|
\leq
\frac{\gamma(1-\lambda)}{1-\gamma\lambda}\epsilon.
\label{eq:app_step_future_bound}
\end{equation}

\paragraph{Chunked advantage.}
For a chunk beginning at $t_0$, ACPPO uses
\begin{equation*}
\hat A_{t_0}^{\mathrm{chunk}}
=
\sum_{j=0}^{h-1}\gamma^j\delta_{t_0+j}
+
\gamma^h\hat A_{t_0+h}^{\mathrm{GAE}},
\end{equation*}
with oracle counterpart
\begin{equation*}
\tilde A_{t_0}^{\mathrm{chunk}}
=
\sum_{j=0}^{h-1}\gamma^j\tilde\delta_{t_0+j}
+
\gamma^h\tilde A_{t_0+h}^{\mathrm{GAE}}.
\end{equation*}
Subtracting the two estimators gives
\begin{align}
\hat A_{t_0}^{\mathrm{chunk}}-
\tilde A_{t_0}^{\mathrm{chunk}}
&=
\sum_{j=0}^{h-1}\gamma^j
\left(\gamma e(s_{t_0+j+1})-e(s_{t_0+j})\right)
+
\gamma^h
\left(
\hat A_{t_0+h}^{\mathrm{GAE}}-
\tilde A_{t_0+h}^{\mathrm{GAE}}
\right).
\label{eq:app_chunk_error_before_cancel}
\end{align}
The first term telescopes:
\begin{equation*}
\sum_{j=0}^{h-1}\gamma^j
\left(\gamma e(s_{t_0+j+1})-e(s_{t_0+j})\right)
=
\gamma^h e(s_{t_0+h})-e(s_{t_0}).
\label{eq:app_chunk_telescoping}
\end{equation*}
Using Eq.~\ref{eq:app_step_gae_error} at time $t_0+h$,
\begin{equation*}
\hat A_{t_0+h}^{\mathrm{GAE}}-
\tilde A_{t_0+h}^{\mathrm{GAE}}
=
-e(s_{t_0+h})+F_{t_0+h}.
\end{equation*}
Substituting this expression into Eq.~\ref{eq:app_chunk_error_before_cancel} yields
\begin{align*}
\hat A_{t_0}^{\mathrm{chunk}}-
\tilde A_{t_0}^{\mathrm{chunk}}
&=
\left(\gamma^h e(s_{t_0+h})-e(s_{t_0})\right)
+
\gamma^h\left(-e(s_{t_0+h})+F_{t_0+h}\right) \\
&=
-e(s_{t_0})+\gamma^h F_{t_0+h}.
\label{eq:app_chunk_error_final}
\end{align*}
The intermediate value errors $e(s_{t_0+1}),\ldots,e(s_{t_0+h})$ cancel. If $\|e\|_\infty\leq\epsilon$, then
\begin{equation*}
\left|\gamma^h F_{t_0+h}\right|
\leq
\frac{\gamma^{h+1}(1-\lambda)}{1-\gamma\lambda}\epsilon.
\label{eq:app_chunk_future_bound}
\end{equation*}
Compared with the recursive term in stepwise GAE in Eq.~\ref{eq:app_step_future_bound}, the remaining recursive bootstrap-error contribution is discounted by an additional factor of $\gamma^h$.

\paragraph{Tradeoff.}
This cancellation does not imply that the chunked advantage is uniformly better. Inside the first $h$ steps, the chunked estimator uses weights $\gamma^j$ rather than the stepwise GAE weights $(\gamma\lambda)^j$. For $\lambda<1$, these weights are larger, making the estimator closer to an $h$-step Monte Carlo target within the chunk. This can increase variance when rewards or transitions are noisy, especially for larger $h$. In addition, all actions in the chunk receive the same chunk-level advantage, which coarsens within-chunk credit assignment. These tradeoffs motivate using short chunks and comparing chunked/stepwise advantages.

\section{Chunked Importance Sampling Ratio Clipping}
\label{app:ratio_diagnostics}

\paragraph{Chunked importance sampling ratio.}
The ACPPO chunked importance ratio can be written as
\begin{equation*}
\rho_{t_0}^{(h)}(\theta)
=
\frac{
\pi_\theta^{(h)}(\mathbf a_{t_0:t_0+h-1}\mid s_{t_0})
}{
\pi_{\theta_{\mathrm{old}}}^{(h)}(\mathbf a_{t_0:t_0+h-1}\mid s_{t_0})
}
=
\prod_{k\in \mathcal T_{t_0}}
\frac{
\pi_{\theta,k}^{(h)}(a_{t_0+k}\mid s_{t_0})
}{
\pi_{\theta_{\mathrm{old}},k}^{(h)}(a_{t_0+k}\mid s_{t_0})
},
\label{eq:app_acppo_ratio_product}
\end{equation*}
where $\pi_{\theta,k}^{(h)}$ denotes the marginal Gaussian for the $k$-th action in the chunk and $\mathcal T_{t_0}$ is the valid prefix of the chunk before any mid-chunk termination.
PPO clipping is applied to the total chunked importance sampling ratio $\rho_{t_0}^{(h)}(\theta)$, not independently to each factor.

For ACPPO-Corr, the corrector observes the current state inside the chunk, so the likelihood factorizes along the realized closed-loop trajectory:
\begin{equation*}
\rho_{t_0}^{\mathrm{corr}}(\theta)
=
\prod_{k\in \mathcal T_{t_0}}
\frac{
\pi_\theta(a_{t_0+k}\mid s_{t_0+k},s_{t_0},k)
}{
\pi_{\theta_{\mathrm{old}}}(a_{t_0+k}\mid s_{t_0+k},s_{t_0},k)
},
\label{eq:app_acppo_corr_ratio_product}
\end{equation*}
If the episode terminates inside a chunk, the product is truncated at the termination boundary and the remaining steps are masked from the regularizers.

\paragraph{Stability of the chunked importance sampling ratio.}
Although the chunk ratio is a product of per-step likelihood ratios, its log-ratio is a sum:
\begin{equation*}
\log \rho_{t_0}^{(h)}
=
\sum_{k=0}^{h-1}
\ell_{t_0+k},
\qquad
\ell_{t}
=
\log \pi_\theta(a_t\mid \cdot)
-
\log \pi_{\theta_{\mathrm{old}}}(a_t\mid \cdot).
\end{equation*}
For intuition, consider Gaussian policies with fixed covariance and small policy updates.
The mean and the variance can be formulated as the KL divergence between likelihoods:
\begin{equation*}
\mathbb{E}[\ell_t]
=
-\mathrm{KL}\!\left(
\pi_{\theta_{\mathrm{old}}}(\cdot\mid\cdot)
\,\|\, 
\pi_\theta(\cdot\mid\cdot)
\right),
\qquad
\mathrm{Var}(\ell_t)
=
2\,\mathrm{KL}\!\left(
\pi_{\theta_{\mathrm{old}}}(\cdot\mid\cdot)
\,\|\, 
\pi_\theta(\cdot\mid\cdot)
\right).
\end{equation*}
The variability of the chunk log-ratio is controlled by the accumulated KL over the chunk:
\begin{equation*}
\mathrm{Var}\!\left[\log \rho_{t_0}^{(h)}\right]
\approx
2
\sum_{k=0}^{h-1}
\mathrm{KL}_{t_0+k}.
\end{equation*}

Following common PPO implementations, we adapt the learning rate based on the measured KL divergence:
\begin{equation*}
\eta
\leftarrow
\begin{cases}
\eta / 1.5, & \text{if } \frac{1}{h}\sum_{k=0}^{h-1}\mathrm{KL}_{t_0+k} > 2\kappa, \\[2mm]
1.5\eta, & \text{if } \frac{1}{h}\sum_{k=0}^{h-1}\mathrm{KL}_{t_0+k} < 0.5\kappa, \\[2mm]
\eta, & \text{otherwise}.
\end{cases}
\end{equation*}
where $\eta$ is the learning rate and $\kappa$ is the KL threshold.
Since the KL is controlled during PPO updates and our chunks are short, the chunked importance sampling ratio remains stable in practice.
The clip fractions per chunk length are displayed in Table~\ref{tab:clipping}.
For the main setting \(h=4\), ACPPO-Corr has a clip fraction of \(0.369\), close to PPO's \(0.365\). 
The clip fraction increases for \(h=8\), consistent with a larger accumulated chunk KL, but remains stable in our experiments.

\begin{table}[t]
    \centering
    \caption{Chunk-level clipping diagnostics. }
    \label{tab:clipping}
    \small
    \begin{tabular}{lcc}
        \toprule
        \textbf{Method} & \textbf{Chunk length} & \textbf{Clip fraction} \\
        \midrule
        PPO & -- & 0.365 \\
        ACPPO-Corr & 2 & 0.364 \\
        ACPPO-Corr & 4 & 0.369 \\
        ACPPO-Corr & 8 & 0.408 \\
        \bottomrule
    \end{tabular}
\end{table}

\end{document}